# Investigating the Privacy Risk of Using Robot Vacuum Cleaners in Smart Environments

Benjamin Ulsmåg[1], Jia-Chun Lin[2] and Ming-Chang Lee[3]

[1,2,3]*Department of Information Security and Communication Technology, Norwegian University of Science and Technology,*
*Gjøvik, Norway*

[1] *benjamin.ulsmag@gmail.com*
[2]*jia-chun.lin@ntnu.no*
[3] *mingchang1109@gmail.com*

26th July 2024

# Investigating the Privacy Risk of Using Robot Vacuum Cleaners in Smart Environments

Benjamin Ulsmåg, Jia-Chun Lin[0000−0003−3374−8536], and Ming-Chang Lee[0000−0003−2484−4366]

Department of Information Security and Communication Technology, Norwegian University of Science and Technology (NTNU), Gjøvik, Norway
{benjamin.ulsmag,mingchang1109}@gmail.com, jia-chun.lin@ntnu.no

**Abstract.** Robot vacuum cleaners have become increasingly popular and are widely used in various smart environments. To improve user convenience, manufacturers also introduced smartphone applications that enable users to customize cleaning settings or access information about their robot vacuum cleaners. While this integration enhances the interaction between users and their robot vacuum cleaners, it results in potential privacy concerns because users' personal information may be exposed. To address these concerns, end-to-end encryption is implemented between the application, cloud service, and robot vacuum cleaners to secure the exchanged information. Nevertheless, network header metadata remains unencrypted and it is still vulnerable to network eavesdropping. In this paper, we investigate the potential risk of private information exposure through such metadata. A popular robot vacuum cleaner was deployed in a real smart environment where passive network eavesdropping was conducted during several selected cleaning events. Our extensive analysis, based on Association Rule Learning, demonstrates that it is feasible to identify certain events using only the captured Internet traffic metadata, thereby potentially exposing private user information and raising privacy concerns.

## 1 Introduction

The use of Internet of Things (IoT) devices and the adoption of smart environments have grown in recent years and are expected to continue expanding [9]. Robot vacuum cleaners, smart lighting, intelligent door locks, and air quality sensors are now common devices in a smart environment. These devices aim to simplify daily tasks and routines for users by automating mundane activities, enhancing comfort, and improving the overall quality of life through increased efficiency and personalized settings.

Robot vacuum cleaners, in particular, have gained popularity in smart environments [11]. They offer the capability to autonomously navigate and clean floors, learning and adapting to the layout of the space over time. Users can personalize their operation through settings and preferences, such as scheduling cleanings or indicating no-go zones, which are often managed through intuitive smartphone applications. Furthermore, the advanced integration of these vacuum cleaners with an ecosystem of other IoT devices in the home significantly enriches their functionality. For example, a robot vacuum cleaner can be programmed to commence its cleaning cycle when connected door locks indicate the user has left the house. Additionally, these smart devices can communicate with each other to optimize energy use and cleaning schedules based on daily usage patterns and real-time environmental data. For instance, the vacuum cleaner could delay its start time if the smart lighting system detects continued activity in a particular area, or it could prioritize cleaning in high-traffic zones during periods of minimal activity, thereby enhancing both efficiency and convenience.

While researchers have investigated the security of robot vacuum cleaners through methods such as penetration tests, vulnerability assessments, and active network eavesdropping, the extent of passive eavesdropping in smart environments where these devices are deployed has not been extensively explored. Passive eavesdropping involves the silent monitoring of data traffic between the vacuum cleaners, their cloud services, and other interconnected smart home devices. This method could reveal how these devices manage sensitive data, their interactions within a smart home network, and the potential exposure of user habits or private information.

Therefore, the objective of this paper is to explore the risk of private information exposure in a smart environment through the analysis of network traffic metadata associated with a robot vacuum cleaner. Our approach adopts the perspective of an attacker, utilizing passive eavesdropping without modifying or interacting with the data transmission. We selected a robot vacuum cleaner from a well-known brand, and deployed it in a real smart environment. Several cleaning events were chosen and individually triggered within this environment multiple times. The corresponding network traffic was collected, and a systematic analysis was conducted to identify unique traffic patterns and signatures associated with each event using Association Rule Learning [4], which is a rule-based machine learning method used to discover interesting relationships and patterns between variables in datasets.

Our analysis revealed that each event could be identified using captured network traffic metadata, specifically through the first few packet sizes extracted from the respective filtered traffic files associated with the event. We attempted to identify both a strict and a less strict signature for each event, and then evaluated the effectiveness of each signature in identifying events through a series of tests in a completely different smart environment. Our findings suggest that it is possible to identify certain events using these signatures, thereby potentially uncovering user habits or routines.

The rest of the paper is organized as follows: Section 2 presents related work. Section 3 describes the methodology used for our investigation. Section 4 details the analysis and identified signature(s) for each event, and Section 5 discusses the results of our evaluation. Finally, Section 6 concludes the paper and outlines future work.

## 2   Related Work

The proliferation of IoT devices in smart environments has raised security concerns. According to Alferidah and Jhanjhi in [5] and Swessi and Idoudi in [18], these issues present across various layers of IoT systems, including hardware, software, and communication. Additionally, the nature of data sharing in smart environments introduces privacy concerns. Gu et al. in [12] conducted a study focused on the analysis of wireless Zigbee traffic within the context of a smart office environment. By passively eavesdropping on wireless traffic, they identified a total of 35 distinct events occurring within the office's smart infrastructure. In addition, they successfully extracted and uncovered private information related to office routines from the traffic data. Alyami et al. in [6] proposed a method designed to capture out-of-network encrypted Wi-Fi traffic. Their method was specifically aimed at distinguishing between different IoT devices within a smart environment. Building on similar concerns regarding security and device identification, Acar et al. in [3] employed machine learning techniques to further refine the process of identifying IoT devices and cataloging their specific actions. Their research extended across a variety of communication protocols, including Wi-Fi, Zigbee, and Bluetooth, which are commonly used by IoT devices for connectivity. The authors also recommended countermeasures to protect these devices against passive eavesdropping attacks.

Sami et al. [17] conducted research on the eavesdropping of private information using laser sensor data from a robot vacuum cleaner. They extracted this sensor data via a side-channel attack targeting the vacuum cleaner. Their study demonstrated the ability to sense object vibrations, such as those in pager bags, and even detect spoken words within the environment. Furthermore, by capturing vibrations from objects like television or music speakers, they could identify specific songs and TV shows with high accuracy. Based on their findings, Sami et al. recommended that manufacturers implement security measures to prevent the extraction of high-precision private data from these devices. Ullrich et al. [19] assessed the communication and security aspects concerning the cloud service and application of a robot vacuum cleaner produced by Neato. The authors identified significant privacy risks due to weak cryptography and shared private keys among devices. The researchers were able to find out personal details about the users from the data, such as their daily schedules, the size of their homes, whether they have pets, and how many people live in their households. Sundström and Nilsson [10] conducted an assessment of security implementations and vulnerabilities associated with the Roborock S7 robot vacuum cleaner, excluding the cloud service security. Their findings indicated that the vacuum cleaner was reasonably secure, but they identified a vulnerability related to Dynamic Host Configuration Protocol (DHCP) starvation attacks from rogue devices on the same network. To mitigate this risk, they recommended basic authentication for networks that control Roborock devices.

While a lot of the past work has focused on finding and exploiting security holes to see how they might affect privacy, our study looks at how user private information could be exposed through passive eavesdropping. Our goal is to understand the real-world risks associated with the use of robot vacuum cleaners. Through this investigation, we aim to uncover the privacy issues that come with robot vacuum cleaners and highlight the critical need for advanced protections to safeguard user privacy.

## 3   Methodology

To investigate the potential for private information exposure via robot vacuum cleaners, we detail our methodology in this section.

### 3.1   Target robot vacuum cleaner selection

To choose a robot vacuum cleaner for our study, we conducted a survey considering different brands, including iRobot, Roborock, Neatsvor, Ecovacs, and iLife. Each of them offers a variety of models to meet diverse customers' needs. We selected the iRobot Roomba i7, as shown in Fig. 1, as our target in this study. This decision was influenced by its popularity [1, 8] and the recommendation for this model in several review articles [13, 15], which highlighted its diverse features and reasonable price at the time of our research in 2023. The analysis results presented in this paper might also be applicable to other vacuum cleaners that share overlapping functionalities with the Roomba i7.

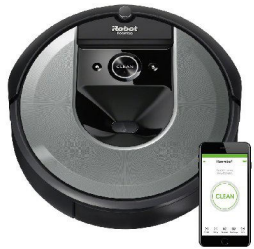

**Fig. 1.** The iRobot Roomba i7.

### 3.2   Smart environment setup

We conducted our study in a real smart environment located in Oslo, Norway. The layout of this environment is shown in Fig. 2. In this environment, we deployed the chosen robot vacuum cleaner, the iRobot Roomba i7, and reset it to its factory default settings. The Oslo environment was equipped with Internet access provided by an external Internet Service Provider (ISP). This setup enabled the robot vacuum cleaner to access to iRobot cloud services and allowed the user to control the device via the iRobot application. To facilitate traffic eavesdropping, a wired local area network (LAN) was established using a LAN switch, and an additional access point (AP) was installed. This AP utilized Network Address Translation (NAT) to consolidate all Wi-Fi traffic from the vacuum cleaner into a single IP address within the smart environment's LAN. This setup simulates a Wide Area Network (WAN) interface, presenting all traffic from the vacuum cleaner as originating from a single address in the smart environment. The AP then directed the traffic to the ISP router through the LAN switch. Simultaneously, the traffic moving through the LAN switch was replicated and forwarded to a Raspberry Pi 3B+ device, which ran the Kali Linux operating system and served as a traffic capturing platform. Several network traffic analysis and capturing tools are included in the Kali Linux distribution. A separate Wi-Fi adapter, the TP-LINK TL-WN722N V2/V3, was acquired and configured in monitor mode to serve as the monitoring wireless NIC. The network infrastructure depicted in Fig. 3 illustrates how the eavesdropping process was carried out, and Table 1 lists all the devices used in this study.

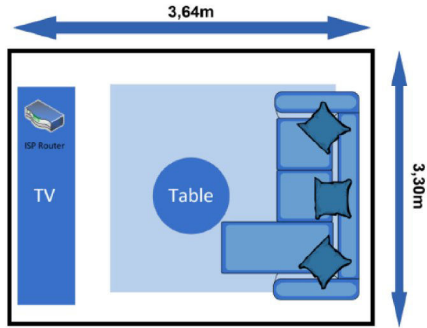

**Fig. 2.** The smart environment in Oslo, Norway.

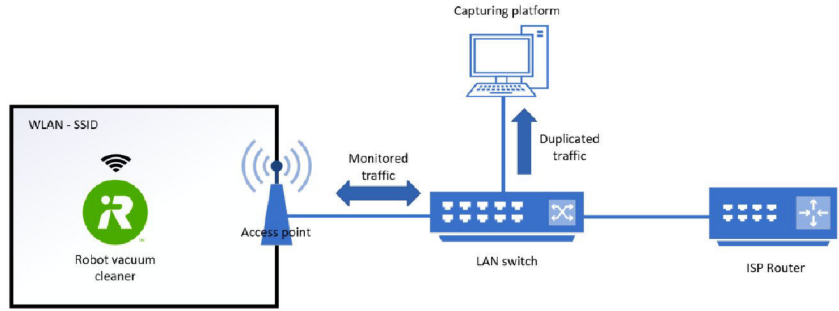

**Fig. 3.** The network infrastructure for passive eavesdropping.

### 3.3 Event selection

We selected six common events with the aim of identifying a unique signature for each, based on their frequency of occurrence and potential security implications. This approach allows us to thoroughly analyze how typical user interactions with robot vacuum cleaners could inadvertently reveal sensitive user information. These six events are listed below.

**Table 1.** Details of all the devices used in our smart environment.

| Device | Details |
| --- | --- |
| Capturing platform | Raspberry Pi 3B+ with Kali Linux |
| Analysis platform | HP Elitebook with Windows 11 |
| Access point (AP) | TP-Link archer MR 200, version 5.30 |
| LAN switch | Cisco catalyst 2960 series 8 port |
| ISP router | Sagemcom Telia |

1. Automated cleaning: This event is triggered through integration with third-party IoT systems or other smart devices. For example, cleaning can be automatically initiated when a user's phone exits their house. Detecting this event could indicate whether the user is away from home, potentially exposing their routines or enabling malicious tracking.

2. App-triggered cleaning: This event is initiated when a user starts cleaning using their iRobot smartphone application. The event finishes when a "finished cleaning" notification is received from the application. Attributing this event could reveal details about the user's smartphone usage and daily routines

3. Scheduled cleaning: This event is initiated based on a user-defined schedule that specifies the cleaning area and start time. Identifying this event could reveal the user's routines or infer whether they are at home during the scheduled time, as this type of cleaning is often planned for times when individuals are typically away from home. This data could potentially be used to ascertain patterns in a user's daily activities, offering insights into their lifestyle and potentially compromising their privacy.

4. Physical-triggered cleaning: This event occurs when the user presses the "Clean" button on the iRobot vacuum cleaner, which triggers the device to perform a full-area cleaning job. If the environment is unfamiliar, this also initiates a mapping process. Detecting this event signals the user's presence within the smart environment, indicating active interaction with the device.

5. App engagement: This event occurs when the user interacts with the iRobot application on their smartphone. Detecting this event reveals user engagement with the iRobot application.

6. Bin removal: This event occurs when the vacuum cleaner's bin is removed, usually after a cleaning cycle or when a notification is sent to the user. It indicates the user's presence alongside the cleaner, potentially disclosing their presence within the smart environment.

In the following sections, we will detail how we collect network traffic for each of these six events and analyze the traffic to identify possible signatures for event identification.

### 3.4   Traffic capturing

To capture traffic for each event, we executed two TShark processes on the Raspberry Pi device within the Oslo environment. One process captured WAN traffic on the Ethernet NIC (eth0), and the other process captured WLAN traffic on the Wireless NIC (wlan1). Note that TShark is the command-line interface version of Wireshark [2, 7], a widely used network protocol analyzer known for its extensive features in capturing and analyzing network traffic.

Before triggering each selected event, we operated the vacuum cleaner in the Oslo environment for one month, ensuring that the traffic we collect was generated during the vacuum cleaner's operational state rather than during the setup phase. Afterwards, we conducted continuous traffic capture for 14 days, during which there was no physical or application interaction. This traffic is referred to as standby traffic in this paper. In our analysis of the standby traffic, we found that approximately 49.2% of the network traffic was related to the Domain Name System (DNS) protocol. Additionally, 26.2% of the traffic was the Transmission Control protocol (TCP), with the majority being the TLS (Transport Layer Security) protocol [16] used by the iRobot Roomba i7 for end-to-end encryption with the cloud server. The last identified protocol was the Address Resolution Protocol (ARP). We filtered out all DNS traffic generated by the AP and TCP-keep-alive traffic since they were irrelevant. Similarly, traffic related to ARP, Dynamic Host Configuration Protocol (DHCP) and Network Time Protocol (NTP) was also excluded for the same reason. After applying this filtering, the total number of packets for the standby traffic dramatically reduced from 5,052,284 to 4,010 (i.e., only 0.8% of the traffic remained).

For each of the six selected events, we triggered the event, recorded the corresponding traffic, and stored the traffic in a single file. This process was repeated 10 times for each event to ensure a sufficient dataset for further analysis.

### 3.5   Traffic analysis and signature identification

To analyze network traffic and identify potential signatures for each event, we adopted a systematic approach, consisting of two phases: Protocol Identification and Signature Identification.

In the first phase, all traffic files associated to each event were imported into Wireshark and analyzed with the supported protocol hierarchy tool, which displays the various protocols and their distributions across the files. We then analyzed the traffic for each identified protocol and filtered out irrelevant traffic by using the same filtering that we used for the standby traffic. We found that all event-specific packets were larger than 97 bytes.

Subsequently, in the second phase, we analyzed the remaining traffic to identify unique signatures for event identification. This involved searching for unique packets for each event, calculating the total number of packets for each event, extracting the first few packet sizes as a sequence from each event, and discovering association between different packet sizes within the sequence using Association Rule Learning (ARL) [4], which is a rule-based machine learning method used to identify common associations or relationships among a set of items in a dataset. It is widely employed to discover which items tend to co-occur, as demonstrated in [14]. In this method, the rule $X \rightarrow Y$ implies that whenever $X$ occurs, $Y$ is likely to occur as well. The condition that the rule holds with minSupport $s$ means that the rule is considered significant if at least $s \cdot 100\%$ of the transactions in the dataset contain both $X$ and $Y$, $0 \leqslant s \leqslant 1$. Note that minSupport represents the minimum frequency or proportion that a set of items (or an association rule) must appear in the dataset to be considered significant. For example, if $s = 0.9$ (or 90%), a rule must appear in at least 90% of all transactions to meet the minimum support criterion. The primary purpose of setting a minSupport threshold is to reduce the number of rules generated by eliminating those that are too rare. This not only focuses on potentially valuable insights but also significantly reduces computational complexity by limiting the number of rules that need evaluation.

## 4   Analysis results and identified signatures

In this section, we present our analysis result and identified signature(s) for each event individually. Additionally, we examine all identified signatures across all events to determine a unique signature for each event.

### 4.1   Automated cleaning

Our analysis of protocol distribution revealed that the following two DNS response packets appeared in all ten automated-cleaning traffic files within the Oslo environment. This consistency suggests that these two packets could serve as a promising signature for identifying the automated cleaning event.

– A DNS response for "0550315.ingest.sentry.io".
– A DNS response for "s3.amazonaws.com".

In addition, we evaluated whether the total number of packets during the automated cleaning event could serve as a signature or not. However, the average total number of packets for the ten automated cleanings was 2425.4 packets, with a standard deviation of 767.27 packets. Due to the high standard deviation, the total number of packets should not be considered as a signature.

Finally, we extracted the first 20 packet sizes from each of the ten filtered automated-cleaning traffic files. We then employed ARL to identify associated packet sizes. In order to discover strong associations rather than explore a wide range of potential relationships, we configured ARL with minSupport at 0.99, minConfidence at 1, and verbosity at 1. Note that minConfidence sets the threshold for how often a rule must be true, and verbosity controls the detail level of the output generated by the analysis.

Fig. 4 displays the analysis results, with each row representing the first 20 packet sizes of an automated cleaning event after all irrelevant traffic was removed. The labels 'D' and 'S' respectively denote that the iRobot Roomba is the destination and source of the packet. The analysis reveals a consistent pattern throughout the ten automated cleaning events: six packet sizes [S175, S176, S179, S446, D1100, D1106] were always found together in every automated cleaning event, regardless of their sequence or frequency of occurrence. Therefore, we considered these six packet sizes as a definitive signature for identifying the automated cleaning event.

Furthermore, in addition to the above strict signature, we attempted to find another smaller subset of packet sizes that might serve as an alternative signature. However, all the results generated by ARL yielded the same set of the six packet sizes, so no less strict signature was found.

```
 1: D289 D316 D316 S176 S187 D409 D404 S175 S480 D1140 S179 S440 D1100 S179 S446 D1106 S176 S475 S179 S253
 2: D510 S176 S187 D409 D271 S179 S440 D1100 S175 S405 D988 S179 S446 D1106 S176 S342 S179 S253 D626 S179
 3: D315 D288 S176 D408 D271 D271 S175 S405 D988 S179 S440 D1100 S179 S446 D1106 S176 S342 S179 S253
 4: D315 D288 S176 S186 D408 D404 S175 S480 D1140 S179 S440 D1100 S179 S446 D1106 S176 S475 S179 S253 D626
 5: D316 D289 S176 S187 D409 D271 S175 S405 D988 S179 S440 D1100 S179 S446 D1106 S176 S342 S179 S253 D626
 6: D316 D289 S176 S187 D409 D271 S175 S405 D988 S179 S440 D1100 S179 S446 D1106 S176 S342 S179 S253 D626
 7: S172 S179 D392 D315 D288 S176 S186 D408 D271 S179 S440 D1100 S175 S405 D988 S179 S446 D1106 S176 S342
 8: S172 S179 D392 D315 D288 S176 S186 D408 D271 S179 S440 D1100 S175 S405 D988 S179 S446 D1106 S176 S342
 9: D289 D316 S176 S187 D409 D271 S179 S440 D1100 S175 S405 D988 S179 S446 D1106 S176 S342 S176 S483 S179
10: S172 S291 D859 D288 D315 S176 S186 D408 D271 S175 S405 D988 S179 S440 D1100 S176 S446 D1106 S176 S342
```

**Fig. 4.** Extraction of the first 20 packet sizes from each of the ten automated cleaning events performed in the Oslo environment. The strict signature we identified, representing the consistent pattern found in all events, is highlighted in grey.

### 4.2   App-triggered cleaning

App-triggered cleaning is initiated when the user needs additional cleaning outside of the pre-scheduled one, which could happen when the user is away from home. Identification of this event can therefore expose private information about user location.

Similar to our analysis of the automated cleaning event, we discovered a consistent occurrence of the two identical DNS response packets in each traffic file associated with the App-triggered cleaning event. This consistency suggests a potential signature for identifying the App-triggered cleaning event. However, we found that the total number of packets of an App-triggered cleaning event was not a reliable indicator for event identification, given the high standard deviation observed.

Our analysis of the occurrence of packet sizes demonstrated potential. We employed ARL on the first 20 packet sizes extracted from each of the ten filtered App-triggered cleaning events, using the same parameter setting as those used for the automated cleaning (i.e., minSupport=0.99, minConfidence=1, and verbosity=1). The analysis shows that three specific packet sizes [S176, S179, D1239] consistently appear together in all the ten events, as illustrated in Fig. 5, independence of their order or how often they occur. Hence, these three packet sizes are considered as a promising signature for the App-triggered cleaning event. In addition, according to the analysis results from ARL, we found that the following three sets of packet sizes could serve as alternative, less strict signatures.

- [S176, D1239]
- [S179, D1239]
- [S176, S179]

However, we omitted the last rule because both S176 and S179 were already included in the signature identified for the automated cleaning, meaning that they cannot uniquely distinguish the App-triggered cleaning event.

**Fig. 5.** Extraction of the first 20 packet sizes from each of the ten App-triggered cleaning events performed in the Oslo environment. The strict signature we identified is highlighted in grey.

### 4.3   Scheduled cleaning

Recall that the scheduled cleaning event is initiated based on a user-defined schedule, specifying both the cleaning area and start time. Our protocol distribution analysis for the scheduled cleaning event reveals a similarity to those for the automated cleaning and App-triggered cleaning events. We observed the same two DNS responses across all 10 scheduled cleaning events within the Oslo environment.

Fig. 6 displays the first 20 packet sizes extracted from each of the ten filtered traffic files related to the scheduled cleaning event. The analysis results from ARL reveal that six specific packet sizes [S176, S179, S253, S448, D626, D1108] are found consistently present across all the ten files, regardless of their sequence or frequency of occurrence. Therefore, we consider this set of packet sizes as a signature for identifying the scheduled cleaning event. It is important to note that we were unable to identify another smaller set of packet sizes as an alternative signature because ARL consistently grouped these six packet sizes together.

**Fig. 6.** Extraction of the first 20 packet sizes from each of the ten scheduled cleaning events in the Oslo environment. The strict signature we identified is highlighted in grey.

### 4.4 Physical-triggered cleaning

When the ten physical-triggered cleaning events were individually performed in the Oslo environment, we also observed the same two DNS responses across all the events. Furthermore, the ARL analysis on the first 20 packet sizes extracted from each filtered traffic file related to the physical-triggered cleaning event shows that nine specific packet sizes [S175, S176, S179, D626, D903, S253, S290, S369, D1106] were consistently observed together (please see Fig. 7). Hence, we consider this set as a promising signature for identification of the physical-triggered cleaning event. However, a less strict signature could not be established due to the strong association among these nine packet sizes.

| # | | | | | | | | | | | | | | | | | | | | |
|---|---|---|---|---|---|---|---|---|---|---|---|---|---|---|---|---|---|---|---|---|
| 1: | S179 | S440 | D1100 | S179 | S448 | D1106 | S176 | S475 | S175 | S369 | D903 | S176 | S290 | S179 | S253 | D626 | S179 | S446 | D1106 | S179 |
| 2: | S179 | S159 | D345 | S179 | S446 | D1106 | S176 | S290 | S175 | S369 | D903 | S176 | S290 | S179 | S253 | D626 | S172 | S179 | D392 | S179 |
| 3: | S179 | S160 | D346 | S179 | S440 | D1100 | S179 | S446 | D1106 | S175 | S369 | D903 | S176 | S290 | S176 | S290 | S179 | S253 | D626 | S179 |
| 4: | S179 | S160 | D345 | S179 | S440 | D1100 | S179 | S446 | D1106 | S175 | S369 | D903 | S176 | S290 | S176 | S290 | S179 | S253 | D626 | S179 |
| 5: | S179 | S440 | D1100 | S179 | S446 | D1106 | S176 | S290 | S175 | S369 | D903 | S176 | S290 | S179 | S253 | D626 | S179 | S446 | D1106 | S179 |
| 6: | S179 | S440 | D1100 | S179 | S446 | D1106 | S175 | S369 | S369 | D903 | S176 | S290 | S176 | S290 | S179 | S253 | D626 | S179 | D1106 | S179 |
| 7: | S172 | S179 | D392 | S179 | S446 | D1106 | S176 | S290 | S175 | S369 | D903 | S176 | S290 | S179 | S253 | D626 | S179 | S446 | D1106 | S179 |
| 8: | S179 | S440 | D1100 | S179 | S446 | D1106 | S176 | S290 | S175 | S369 | D903 | S176 | S290 | S179 | S253 | D626 | S179 | S446 | D1106 | S179 |
| 9: | S179 | S440 | D1100 | S179 | S448 | D1106 | S176 | S290 | S175 | S369 | D903 | S172 | S233 | D551 | S176 | S290 | S179 | S253 | D626 | S179 |
| 10: | S179 | S159 | D345 | S179 | D1100 | S179 | S446 | D1106 | S175 | S369 | D903 | S176 | S290 | S179 | S253 | D626 | S179 | | | |

**Fig. 7.** Extraction of the first 20 packet sizes from each of the ten physical-triggered cleaning events in the Oslo environment. The strict signature we identified is highlighted in grey.

### 4.5 App engagement

Recall that an App engagement event is initiated when the user opens and engages with the iRobot application. For our analysis, we activated and interacted with the iRobot application, without focusing on any specific action. Various actions were executed, including changing the scheduled cleaning time, viewing the dashboard, and adjusting settings, etc.

The protocol distribution analysis for this event revealed the presence of only TCP packets; no DNS packets were observed during the event. The requested information pulled from the iRobot Roomba during application engagement was initiated from a2uowfjvhio0fa.iot.useast-1.amazonaws.com.

| # | | | | | | | | | | | | | | | | | | | | |
|---|---|---|---|---|---|---|---|---|---|---|---|---|---|---|---|---|---|---|---|---|
| 1: | D209 | S315 | D288 | S298 | D408 | S176 | S1053 | D1514 | D1514 | D1084 | D1514 | D1514 | D1111 | S174 | S140 | D333 | S175 | S1514 | S569 | D1514 |
| 2: | D208 | D316 | D289 | S176 | S187 | D409 | S176 | S1052 | D1514 | D1514 | D1112 | D1514 | D1514 | D1085 | S174 | S140 | D333 | S175 | S1514 | S570 |
| 3: | D208 | D537 | S176 | S186 | D408 | S176 | S1052 | D1514 | D1514 | D1085 | D1514 | D1514 | D1112 | S174 | S140 | D333 | S175 | S1514 | S570 | D1514 |
| 4: | S179 | S160 | D346 | D208 | D289 | D316 | S176 | S187 | D409 | S176 | D1052 | D1514 | D1514 | D1111 | D1514 | D1514 | D1084 | S174 | S140 | D333 |
| 5: | D209 | S315 | D288 | S176 | S186 | D408 | S176 | S1053 | D1514 | D1514 | D1085 | D1514 | D1514 | D1112 | D1514 | D1514 | D1085 | S174 | S140 | D333 |
| 6: | D205 | D289 | D316 | S176 | S1046 | S176 | S187 | D409 | D1514 | D1514 | D1112 | D1514 | D1514 | D1085 | S174 | S140 | D333 | S175 | S1514 | S570 |
| 7: | D207 | D315 | D288 | S176 | S186 | D408 | S176 | S1051 | D1514 | D1514 | D1085 | D1514 | D1514 | D1112 | S174 | S140 | D333 | S175 | S1514 | S570 |
| 8: | S179 | S159 | D345 | D207 | D289 | D316 | S176 | S187 | D409 | S176 | S1051 | D1514 | D1514 | D1514 | D1514 | D1514 | D654 | S174 | S140 | D333 |
| 9: | D207 | D508 | S176 | S1050 | S176 | S186 | D408 | D1514 | D1514 | D1112 | D1514 | D1514 | S174 | S140 | D333 | S175 | S1514 | S570 | | |
| 10: | D208 | D316 | D289 | S176 | S1052 | S176 | S187 | D409 | S172 | S219 | D505 | D1514 | D1514 | D1085 | D1514 | D1514 | D1112 | S174 | S140 | D333 |

**Fig. 8.** Extraction of the first 20 packet sizes from the ten App engagement events in the Oslo environment. The strict signature we identified is highlighted in grey.

Fig. 8 displays the first 20 packet sizes extracted from each of the ten filtered traffic files related to the App engagement event. Five specific packet sizes [S140, S174, S176, D333, D1514] were consistently observed together in all the ten files, regardless of their sequence or frequency of occurrence. Therefore, this set of packet sizes is considered a signature for recognizing the App engagement event. We also found another less strict signature, consisting of four specific packet sizes [S140, S174, S176, D333]. Hence, this subset is considered as an alternative signature for the App engagement event.

### 4.6   Bin removal

Recall that the bin removal event occurs when the physical bin eject button on the iRobot Roomba i7 is pressed, thereby releasing the bin. This event was individually executed 10 times by us in the Oslo environment. Our observations revealed that only few packets were generated per event. However, it exhibited a high standard deviation. As a result, the total number of packets captured during the event cannot serve as a reliable signature for identifying the bin removal event.

Following our methodology, we extracted the first few packet sizes from each filtered traffic file and employed ARL to find out associated rules. The results reveal that two specific packet sizes [S186, D410] were consistently observed across all the 10 files, as illustrated in Fig. 9. Hence, these two sizes are considered as a signature for recognizing the bin removal event. Given that the signature consists only two packet sizes, we did not pursue any less strict signatures for this event.

| | | | | | | | | | | | | | | | | | | | |
|---|---|---|---|---|---|---|---|---|---|---|---|---|---|---|---|---|---|---|---|
| 1: D208 | D288 | D315 | S176 | **S186** | D408 | S176 | S1052 | S179 | S450 | D1110 | S179 | **S186 D410** | S179 | S450 | D1110 | S179 | S185 | D409 |
| 2: S179 | S448 | D1108 | S179 | S187 | D411 | S179 | S448 | D1108 | S179 | **S186 D410** | | | | | | | | |
| 3: S179 | S160 | D346 | S172 | S233 | D551 | S179 | S450 | D1110 | S179 | S187 | D411 | S179 | S450 | D1110 | S179 | S450 | D1110 | S179 | S450 |
| 4: S179 | S492 | D1222 | S179 | S450 | D1110 | S179 | **S186 D410** | | | | | | | | | | | |
| 5: S179 | S448 | D1108 | S179 | S187 | D411 | S179 | S448 | D1108 | S179 | **S186 D410** | | | | | | | | |
| 6: S603 | D1220 | S179 | S448 | D1108 | S179 | **S186 D410** | | | | | | | | | | | | |
| 7: S179 | S490 | D1220 | S179 | S448 | D1108 | S179 | **S186 D410** | | | | | | | | | | | |
| 8: S325 | D505 | S179 | S448 | D1108 | S179 | S187 | D411 | S179 | S448 | D1108 | S179 | **S186 D410** | S172 | S179 | D392 | | | |
| 9: S179 | S490 | D1220 | S179 | S448 | D1108 | S179 | **S186 D410** | | | | | | | | | | | |
| 10: S179 | S490 | D1220 | S179 | S448 | D1108 | S179 | S448 | D1108 | **S186 D410** | | | | | | | | | |

**Fig. 9.** The first few packet sizes extracted from the ten bin removal events in the Oslo environment. The identified signature is highlighted in grey.

### 4.7   Summary

Based on all the analysis results mentioned above, we confirm that the following two DNS responses were consistently found in all cleaning events, including automated cleaning, App-triggered cleaning, scheduled cleaning, and physical-triggered cleaning events. However, these responses did not appear in other events, such as the App engagement and Bin removal events. Therefore, while these two DNS responses can be used to identify if a cleaning event occurs or not, they cannot be used to identify each individual cleaning event.

– A DNS response for "0550315.ingest.sentry.io".
– A DNS response for "s3.amazonaws.com".

Table 2 summarizes all identified signatures for each event. Apparently, each strict signature is unique to its corresponding event even though there is some slight overlapping between different events. Furthermore, it is worth noting that four out of the six events do not have a less strict signature identified. This is because the corresponding ARL analysis results suggest a strong correlation between the components within the strict signature for these four events.

**Table 2.** All identified signatures for each event.

| Event | Identified signatures |
|---|---|
| Automated cleaning | Strict: [S175, S176, S179, S446, D1100, D1106] <br> Less strict: none |
| App-triggered cleaning | Strict: [S176, S179, D1239] <br> Less strict: [S176, D1239] or [S179, D1239] |
| Scheduled cleaning | Strict: [S176, S179, S253, S448, D626, D1108] <br> Less strict: none |
| Physical-triggered cleaning | Strict: [S175, S176, S179, D626, D903, S253, S290, S369, D1106] <br> Less strict: none |
| App engagement | Strict: [S140, S174, S176, D333, D1514] <br> Less strict: [S140, S174, S176, D333] |
| Bin removal | Strict: [S186, D410] <br> Less strict: none |

## 5   Evaluation

To evaluate the effectiveness of each identified signature in event identification, we conducted a series of tests in another smart environment located in Drammen, Norway. As illustrated in Fig. 10, this environment has a different size and layout as compared with the Oslo environment. Similar to the Oslo environment, we established a wired and wireless network infrastructure, Internet connection, set up the Raspberry Pi device for traffic eavesdropping, etc. This setup allowed us to conduct traffic eavesdropping from this environment.

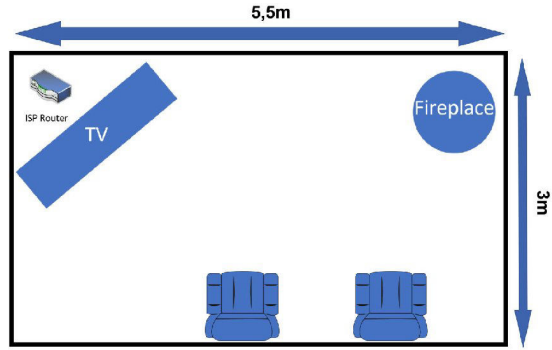

**Fig. 10.** The smart environment in Drammen, Norway.

**Automated cleaning**
```
 1: D315 D288 S176 S186 D408 D404 S175 S425 D982 S179 S439 D1099 S179 S445 D1105 S176 S474 S176 S615 S179
 2: D289 D316 S176 S187 D409 D404 S175 S449 D1046 S179 S440 D1100 S179 S446 D1106 S176 S475 S179 S253 D626
 3: S172 S288 D714 D316 D289 S176 S187 D409 D404 S175 S425 D982 S179 S439 D1099 S179 S445 D1105 S176 S474
 4: D315 D288 S176 S186 D408 D404 S179 S439 D1099 S175 S425 D982 S179 S445 D1105 S176 S615 S179 S253 D626
 5: D289 D316 S176 S187 D409 D404 S175 S410 D936 S179 S446 D1106 S176 S475 S176 S616 S179 S253 D626 S179
 6: D289 D316 D316 S176 S187 D409 D404 S175 S449 D1046 S179 S440 D1100 S179 S446 D1106 S179 S475 S176 S616
 7: D288 D315 S176 S186 D408 D404 S175 S449 D1046 S179 S440 D1100 S179 S446 D1106 S176 S475 S176 S616 S179
 8: D289 D316 S297 D409 D404 S179 S440 D1100 S175 S449 D1046 S179 S446 D1106 S176 S475 S176 S616 S179 S253
 9: S172 S233 D551 D315 D288 S176 S186 D408 D404 S175 S449 D1046 S179 S440 D1100 S179 S446 D1106 S176 S475
10: D289 D316 S176 S187 D409 D404 S179 S440 D1100 S175 S449 D1046 S179 S446 D1106 S176 S475 S176 S616 S179
```

**App-triggered cleaning**
```
 1: D209 D315 D288 S176 S186 D408 S176 S949 D503 S175 S540 D1200 S179 S440 D1100 S179 S446 D1106 S176 S715
 2: D208 D289 D316 S176 S187 D409 S176 S1514 S227 D503 S175 S509 D1106 S179 S440 D1100 S179 S446 D1106 S574
 3: D209 D316 D289 S176 S187 D409 S176 S1514 S1514 S787 D503 S175 S524 D1152 S179 S440 D1100 S179 S446
 4: D209 D288 D315 S176 S1085 S176 S186 D408 D503 S175 S540 D1200 S179 S440 D1100 S179 S446 D1106 S176 S574
 5: S179 S160 D346 D209 D316 D289 S176 S1514 S256 S176 S187 D409 D503 S175 S540 D1200 S179 S440 D1100 S179
 6: S172 S179 D392 D208 D315 D288 S176 S186 D408 S176 S1359 D503 S175 S509 D1106 S179 S440 D1100 S179 S446
 7: D207 D289 D316 S176 S187 D409 S176 S1514 S1514 S1514 S1514 S1514 S205 D318 S175 S398 D869 S179 S440 D1100
 8: D207 D316 D288 S176 S186 D408 S176 S1514 S1514 S1514 S926 D318 S175 S398 D869 S179 S446 D1106
 9: D208 D315 D288 S176 S985 S175 S186 D408 D208 S176 D316 D289 S985 S176 S187 D409 D503 S175 S509 D1106
10: D208 D315 D288 S176 S186 D408 S176 S1514 S255 D503 S175 S509 D1106 S179 S440 D1100 S179 S446 D1106 S176
```

**Scheduled cleaning**
```
 1: S175 S466 D1094 S179 S442 D1102 S179 S448 D1108 S176 S477 S179 S253 D626 S179 S448 D1108 S179 D1108
 2: S175 S466 D1094 S179 S442 D1102 S179 S448 D1108 S176 S618 S179 S253 D626 S176 S835 S179 S448 D1108
 3: S175 S466 D1094 S179 S442 D1102 S179 S448 D1108 S176 S477 S176 S618 S179 S253 D626 S176 S835 S179 S448
 4: S175 S466 D1094 S179 S442 D1102 S179 S448 D1108 S176 S477 S176 S618 S179 S253 D626 S179 S448 D1108 S176
 5: S175 S466 D1094 S179 S442 D1102 S179 S448 D1108 S176 S618 S179 S253 D626 S179 S448 D1108 S176 S835 S179
 6: S175 S466 D1094 S179 S442 D1102 S179 S448 D1108 S176 S477 S176 S618 S179 S253 D626 S176 S835 S179 S448
 7: S175 S466 D1094 S179 S442 D1102 S179 S448 D1108 S176 S477 S176 S618 S179 S253 D626 S179 S448 D1108 S179
 8: S175 S466 D1094 S179 S442 D1102 S179 S448 D1108 S176 S477 S176 S618 S179 S253 D626 S179 S448 D1108 S179
 9: S172 S234 D552 S175 S466 D1094 S179 S442 D1102 S179 S448 D1108 S176 S477 S176 S618 S179 S253 D626 S179
10: S175 S466 D1094 S179 S442 D1102 S179 S448 D1108 S176 S618 S179 S253 D626 S176 S835 S179 S448 D1108 S179
```

**Physical-triggered cleaning**
```
 1: S175 S314 D745 S179 S447 D1107 S179 S447 D1107 S179 S445 D1105 S179 S253 D626 S179 S445 D1105 S179 S445
 2: S179 S446 D1106 S176 S290 S175 S338 D809 S176 S290 S179 S446 D1106 S179 S446 D1106 S179 S446 D1106 S176
 3: S179 S440 D1100 S179 S446 D1106 S176 S290 S175 S338 D809 S176 S290 S179 S253 D626 S179 S446 D1106 S179
 4: S179 S159 D345 S179 S440 D1100 S179 S446 D1106 S175 S338 D809 S176 S290 S176 S290 S179 S179 S253 D626
 5: S179 S160 D346 S179 S439 D1099 S179 S445 D1105 S314 D745 S176 S289 S176 S253 D626 S179 S445 D1105 S179
 6: S179 S439 D1099 S179 S445 D1105 S175 S314 D745 S176 S430 S179 S253 D626 S179 S445 D1105 S179 S445 D1105
 7: S179 S160 D346 S179 S439 D1099 S179 S445 D1105 S175 S314 D745 S176 S289 S176 S253 D626 S179
 8: S172 S233 D551 S179 S439 D1099 S179 S445 D1105 S175 S314 D745 S176 S289 S176 S289 S179 S253 D626 S179
 9: S179 S440 D1106 S179 S446 D1106 S175 S338 D809 S176 S574 S176 S431 S179 S253 D626 S179 S648 S179 S446
10: S179 S440 D1100 S179 S446 D1106 S175 S338 D809 S176 S290 S176 S290 S179 S253 D626 S176 S448 D1106 S176
```

**App engagement**
```
 1: D209 D288 D315 S296 D408 S176 S1514 S376 D879 D852 S174 S140 D333 S175 S469 D904 S175 S346 D848
 2: D209 D289 D316 S176 S187 D409 S176 S1514 S131 D852 D879 S174 S140 D333 S175 S469 D904 S175 S346 D848
 3: D209 D315 D288 S176 S186 D408 S176 S1514 S131 D879 D852 S174 S140 D333 S175 S469 D904 S175 S346 D848
 4: D209 D289 D316 S176 S1514 S159 S176 S187 D409 D852 S174 D879 D825 D852 S174 S140 D333 S175 S346 D848
 5: S172 S179 D392 D208 D315 S176 S186 D408 S176 S987 D825 D852 S174 S140 D333 S175 S466 D877 S175
 6: D208 D289 D316 S176 S187 D409 S176 S1514 S158 D852 D825 S174 S140 D333 S175 S466 D877 S175 S346 D848
 7: D208 D288 D315 S176 S1361 S176 S186 D408 D852 D825 S174 S140 D333 S175 S466 D877 S175
 8: D208 D289 D316 S176 S187 D409 S176 S1514 S98 D851 D824 S174 S140 D333 S175 S466 D876 S175 S346 D848
 9: D208 D288 D315 S176 S1514 S126 S176 S186 D408 D825 D852 S174 S140 D333 S175 S466 D877 S175 S346 D848
10: D208 D289 D316 S176 S187 D409 S176 D1389 D825 D852 S174 S140 D333 S175 S466 D877 S175 S346 D848
```

**Bin removal**
```
 1: S179 S448 D1108 S179 S187 D411
 2: S179 S448 D1108 S179 S187 D411 S179 S448 D1108 S179 S186 D410
 3: S172 S293 D861 S179 S448 D1108 S179 S187 D411 S179 S448 D1108 S179 S186 D410
 4: S179 S448 D1108 S179 S187 D411 S172 S179 D392 S179 S448 D1108 S179 S186 D410
 5: S172 S179 D505 S179 S448 D1108 S179 S187 D411 S172 S234 D552
 6: S179 S448 D1108 S179 S187 D411 S179 D1108 S179 S448 D1108 S179 S489 D1219
 7: S179 S448 D1108 S179 S187 D411 S179 S448 D1108 S179 S186 D410
 8: S561 D1108 S179 S187 D411 S179 S448 D1108 S179 S448 D1108 S179 S186 D410
 9: S179 S448 D1108 S179 S187 D411 S179 S448 D1108 S179 S448 D1108 S179 S186 D410
10: S179 S448 D1108 S179 S187 D411 S179 S448 D1108 S179 S186 D410
```

**Fig. 11.** The first few packet sizes extracted from each of the 60 events triggered in the Drammen environment.

Before deploying the iRobot Roomba i7 to the Drammen environment, we also reset it to its factory default settings. Following the same methodology presented in Section 3, we triggered each of the six events in the Drammen environment, recorded the corresponding network traffic in a single file, and applied the same filter that we used in the Oslo environment to remove any irrelevant traffic. This procedure was individually carried out 10 times for each event in the Drammen environment. Therefore, there are a total of 60 filtered files associated with the six events.

Fig. 11 depicts the first 20 packet sizes extracted from each of these 60 event traffic files after irrelevant traffic has been removed. Therefore, there are 60 lines in this figure. For each signature listed in Table 2, we evaluated how many corresponding events shown in Fig. 11 this signature can accurately recognize. This number is referred to as true positive ($TP$) in this paper. For example, if a signature identified for the scheduled cleaning event in the Oslo environment, denoted as $S$, accurately identifies 8 out of 10 actual scheduled cleaning events in the Drammen environment, then the $TP$ count for signature $S$ is 8. In addition, we evaluated how many matching events in the Drammen environment were not accurately identified by each signature, referred to as false negative ($FN$). Following the previous example, the $FN$ count for signature $S$ is 2. Furthermore, we assess how many mismatching events in the Drammen environment were identified by the signature. This number is referred to as false positive ($FP$). For instance, if $S$ mistakenly identifies 5 events in the Drammen environment as scheduled cleaning events when they actually are not, then the $FP$ count for $S$ is 5.

For each signature, we calculated three widely recognized metrics: Precision, Recall, and F1-score, using the equations below.

$$P = \frac{TP}{TP + FP} \tag{1}$$

$$R = \frac{TP}{TP + FN} \tag{2}$$

$$F1 = 2 \cdot \frac{P \cdot R}{P + R} \tag{3}$$

Precision ($P$ for short) measures the accuracy of the positive predictions made by the signature. Recall ($R$ for short) measures the ability of the signature to identify all relevant instances accurately, and F1-score ($F1$ for short) provides a balanced measure between precision and recall.

Table 3 presents the $TP$, $FP$, and $FN$ results of each signature, whereas Table 4 lists the event identification results of each signature. The results indicate that the signature [S176, S179, S253, S448, D626, D1108] achieves an F1-score of 1, meaning that it successfully identified all the 10 scheduled cleaning events in the Drammen environment without any false identifications. Therefore, this signature represents a reliable identifier for the scheduled cleaning event. Similarly, the signature [S140, S174, S176, D333] also achieves the highest F1-score of 1, making it reliable for identifying the App engagement event. Additionally, by using the signature [S186, D410], we were able to correctly recognize 7 out of the 10 bin removal events in the Drammen environment without introducing any false positive, resulting in an F1-score of 0.824.

**Table 3.** The *TP*, *FP*, and *FN* results of each signature.

| Signature | Associated event | TP | FP | FN |
|---|---|---|---|---|
| [S175, S176, S179, S446, D1100, D1106] | Automated cleaning | 6 | 10 | 4 |
| [S176, S179, D1239] | App-triggered cleaning | 0 | 0 | 10 |
| [S176, D1239] | App-triggered cleaning | 0 | 0 | 10 |
| [S179, D1239] | App-triggered cleaning | 0 | 0 | 10 |
| [S176, S179, S253, S448, D626, D1108] | Scheduled cleaning | 10 | 0 | 0 |
| [S175, S176, S179, D626, D903, S253, S290, S369, D1106] | Physical-triggered | 0 | 0 | 10 |
| [S140, S174, S176, D333, D1514] | App engagement | 0 | 0 | 10 |
| [S140, S174, S176, D333] | App engagement | 10 | 0 | 0 |
| [S186, D410] | Bin removal | 7 | 0 | 3 |

**Table 4.** The event identification results of each signature.

| Signature | Associated event | P | R | F1 |
|---|---|---|---|---|
| [S175, S176, S179, S446, D1100, D1106] | Automated cleaning | 0.375 | 0.6 | 0.462 |
| [S176, S179, D1239] | App-triggered cleaning | 0 | 0 | undefined |
| [S176, D1239] | App-triggered cleaning | 0 | 0 | undefined |
| [S179, D1239] | App-triggered cleaning | 0 | 0 | undefined |
| [S176, S179, S253, S448, D626, D1108] | Scheduled cleaning | 1 | 1 | 1 |
| [S175, S176, S179, D626, D903, S253, S290, S369, D1106] | Physical-triggered | 0 | 0 | undefined |
| [S140, S174, S176, D333, D1514] | App engagement | 0 | 0 | undefined |
| [S140, S174, S176, D333] | App engagement | 1 | 1 | 1 |
| [S186, D410] | Bin removal | 1 | 0.7 | 0.824 |

Notably, five out of the nine identified signatures failed to accurately recognize their associated events within the Drammen environment, yielding a Precision of 0 and a recall of 0. Upon closer inspection, we observed that three signatures are associated with the App-triggered cleaning event, indicating that none of these signatures can serve as a reliable indicator for recognizing the app-triggered cleaning event. Furthermore, we also found that the only signature identified for the physical-triggered cleaning could not recognize any physical-triggered cleaning event occurring in the Drammen environment, making it an unreliable indicator. According to the analysis results, although the App engagement event could not be identified using the strict signature [S140, S174, S176, D333, D1514], it was successfully identified using the less strict signature [S140, S174, S176, D333], achieving an F1-score of 1 without introducing any false positives or false negatives.

The implications of the above results suggest that if any malicious individuals are aware of these reliable signatures, they could determine when a household schedules a cleaning, when a user interacts with their iRobot application on their smartphones, and whether the user is present in their homes to remove the bin from their iRobot Roomba i7. This information enables malicious individuals to infer personal habits, routines, and even times when the user might be away from home, which could then be exploited for harmful intentions.

# 6   Conclusions and Future Work

In this paper, we have investigated the privacy risk associated with the use of robot vacuum cleaners in smart environments, particularly through passive network eavesdropping. Despite the implementation of end-to-end encryption by manufacturers to protect user data, our findings demonstrate that unencrypted network header metadata can still expose private and sensitive information. By deploying a popular robot vacuum cleaner model in a real smart environments, conducting passive eavesdropping, and analyzing traffic through a systematic approach, we were able to identify unique signatures for several cleaning events.

Our experiment conducted in a completely different smart environment demo -nstrated that certain identified signatures can accurately recognize events such as scheduled cleaning, application engagement, and bin removal. This capability implies that malicious individuals could exploit these signatures to further infer user habits and routines. Our study highlights the urgent need for enhanced measures to protect user privacy and advocates for a comprehensive approach to secure robot vacuum cleaners.

In our future work, we plan to develop an automated tool to streamline the capturing, processing, and analysis of network traffic using various machine learning methods. This tool will facilitate our privacy studies on robot vacuum cleaners across different brands and various IoT devices, enabling us to comprehensively assess and address privacy concerns in the rapidly evolving landscape of smart environments.

## Acknowledgement

The authors want to thank the anonymous reviewers for their reviews and valuable suggestions to this paper. This work has received funding from the Research Council of Norway through the SFI Norwegian Centre for Cybersecurity in Critical Sectors (NORCICS) project no. 310105.